\documentclass[10pt,twocolumn,letterpaper]{article}

\usepackage[pagenumbers]{cvpr} 

\usepackage[accsupp]{axessibility}

\usepackage[dvipsnames]{xcolor}

\usepackage{amsmath}
\usepackage{amssymb}
\usepackage{booktabs}
\usepackage{multirow}
\usepackage{epstopdf}
\usepackage{bm}
\usepackage{url}
\usepackage{makecell}
\usepackage{cite}
\usepackage{color}
\usepackage{bbding}
\usepackage{xcolor}
\usepackage{colortbl}
\definecolor{Gray}{gray}{0.95}
\definecolor{Cyan}{rgb}{0.88,1,1}
\usepackage{subcaption}
\definecolor{Grayy}{gray}{0.6}

\usepackage{adjustbox}

\definecolor{mypink}{rgb}{.99,.95,.95}  
\definecolor{myblue}{rgb}{.93,.97,.99}
\definecolor{mygreen}{rgb}{.95,.99,.95}
\definecolor{bestcolor}{gray}{.9}
\definecolor{cvprblue}{rgb}{0.21,0.49,0.74}
\usepackage[pagebackref,breaklinks,colorlinks,citecolor=cvprblue]{hyperref}

\def\confName{CVPR}
\def\confYear{2024}

\title{FineParser: A Fine-grained Spatio-temporal Action Parser for Human-centric Action Quality Assessment}

\author{
Jinglin Xu\textsuperscript{1}\quad
Sibo Yin\textsuperscript{2}\quad
Guohao Zhao\textsuperscript{2}\quad 
Zishuo Wang\textsuperscript{2}\quad 
Yuxin Peng\textsuperscript{2}\thanks{}\\
\textsuperscript{1} School of Intelligence Science and Technology, University of Science and Technology Beijing\\
\textsuperscript{2} Wangxuan Institute of Computer Technology, Peking University\\
{\tt\small xujinglinlove@gmail.com; 2000012982@stu.pku.edu.cn; ssee7235@gmail.com;} \\
\vspace{-6pt}
{\tt\small 1900013093@pku.edu.cn; pengyuxin@pku.edu.cn}
}

\begin{document}

\twocolumn[{
\renewcommand\twocolumn[1][]{#1}
\maketitle
\begin{center}
\vspace{-10mm}
    \centering
    \captionsetup{type=figure}
    \includegraphics[width=0.96\linewidth]{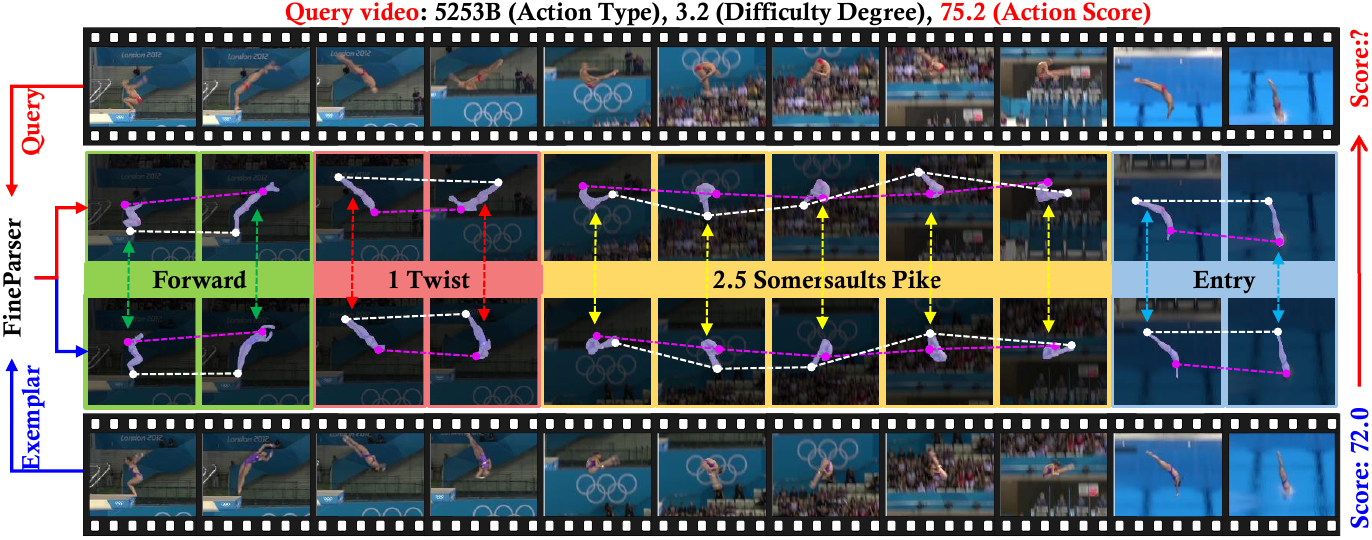}
    \vspace{-10pt}
    \captionof{figure}{An overview of fine-grained spatial-temporal action parser (\textbf{\textit{FineParser}}). It enhances human-centric foreground action representations by exploiting fine-grained semantic consistency and spatial-temporal correlation between video frames, improving the AQA performance. Green, red, yellow, and blue dashed lines represent the fine-grained alignment of target actions between query and exemplar videos in time and space within the same semantics.}
    \label{top}
\end{center}
}]

\footnotetext[1]{Corresponding author.}

\maketitle
\begin{abstract}
\vspace{-7pt}
Existing action quality assessment (AQA) methods mainly learn deep representations at the video level for scoring diverse actions. Due to the lack of a fine-grained understanding of actions in videos, they harshly suffer from low credibility and interpretability, thus insufficient for stringent applications, such as Olympic diving events.
We argue that a fine-grained understanding of actions requires the model to perceive and parse actions in both time and space, which is also the key to the credibility and interpretability of the AQA technique.
Based on this insight, we propose a new fine-grained spatial-temporal action parser named \textbf{FineParser}. It learns human-centric foreground action representations by focusing on target action regions within each frame and exploiting their fine-grained alignments in time and space to minimize the impact of invalid backgrounds during the assessment. In addition, we construct fine-grained annotations of human-centric foreground action masks for the FineDiving dataset, called \textbf{FineDiving-HM}. With refined annotations on diverse target action procedures, FineDiving-HM can promote the development of real-world AQA systems. Through extensive experiments, we demonstrate the effectiveness of FineParser, which outperforms state-of-the-art methods while supporting more tasks of fine-grained action understanding.
Data and code are available at \url{https://github.com/PKU-ICST-MIPL/FineParser_CVPR2024}.

\end{abstract}
\vspace{-9pt}    
\section{Introduction}
\label{sec:intro}

Video understanding is a crucial technique in computer vision that aims to analyze objects, actions, or events in videos automatically. It is essential for many real-world applications, e.g., human-computer interaction~\cite{fieraru2020three,ng2020you2me,hassan2021populating,wang2021synthesizing}, medical rehabilitation~\cite{wang2023areas,gupta2023dataset}, and sports analysis \cite{shao2020finegym,li2021multisports,xu2022finediving,Cui_2023_ICCV}. Notably, a clear and accurate understanding of \textit{actions} in videos provides critical and extensive technique support in action quality assessment (AQA). This considerably impacts sports analysis, helping evaluate athlete performance, designing targeted training programs, and preventing sports injuries.

Unlike general videos, sports videos are sequential processes with explicit procedural knowledge. Athletes have to complete a series of rapid and complex movements. Taking diving as an example, athletes will stretch, curl, and move their limbs and joints to finish different somersaults with three body positions, including straight, pike, and tuck, interspersed with varying twists. Then, the referee will assess the scores based on the athletes' take-off, somersault, twists, and entry. To achieve better competitive performance, athletes (1) take off decisively and forcefully at the right angle and with a proper height; (2) perform beautiful body positions, quick somersaults, and twists in the flight; (3) enter the water with a posture perpendicular to the surface, avoiding splashing water around. According to the diving rules, just a few degree differences in the take-off angle/height and the verticality of entry into the water can affect the number of points deducted. The difficulty lies in whether the human eye can accurately discern such subtle differences.

To address this issue, many video understanding-based AQA methods \cite{parmar2019action, xu2019learning, tang2020uncertainty,yu2021group} lack a fine-grained understanding of actions in videos. They cannot solve the problem of limitations of human eye judgment and lack credibility, which is inadmissible in real-world applications. There is an urgent need for a fine-grained understanding of actions, i.e., parsing the internal structures of actions in time and space with semantic consistency and spatial-temporal correlation, to obtain precise action representations and improve the usefulness of the AQA system.

To this end, we present a new framework for fine-grained action understanding, which learns human-centric foreground action representations with context information by developing a new fine-grained spatial-temporal action parser named \textit{\textbf{FineParser}}. FineParser consists of four components: (1) spatial action parser (SAP); (2) temporal action parser (TAE); (3) static visual encoder (SVE); (4) fine-grained contrastive regression (FineReg).
Given query and exemplar videos, SAP first models the intra-frame feature distribution of each video by capturing multi-scale representations of human-centric foreground actions. The critical regions are concentrated around the athlete's body, springboard (or platform), and splash, guaranteeing the spatial parsing to be credible and visually interpretable.
Then, TAP models semantic and temporal correspondences between videos by learning their spatial-temporal representations and parsing the actions into consecutive steps. Combining TAP and SAP, FineParser learns the target action representations at the fine-grained level, ensuring semantic consistency and spatial-temporal correspondence across videos.
In addition, SVE enhances the above target action representations by capturing more contextual details.
Finally, FineReg can quantify the quality differences in pairwise steps between query and exemplar videos and assess the action quality.

To promote the evaluation of credibility and visual interpretability of FineParser, we densely label human-centric foreground action regions of all videos in the FineDiving dataset and construct additional mask annotations, named \textbf{\textit{FineDiving-HM}}. Experimental results demonstrate that our fine-grained actions understanding framework accurately assesses diving actions while focusing on critical regions consistent with human visual understanding.

The contributions of this paper are summarized as follows:
(1) We propose a new fine-grained spatial-temporal action parser, \textbf{\textit{FineParser}}, beneficial to the AQA task via human-centric fine-grained alignment.
(2) FineParser captures the human-centric foreground action regions within each frame, minimizing the impact of invalid background in AQA.
(3) We provide human-centric foreground action mask annotations for the FineDiving dataset, \textbf{\textit{FineDiving-HM}}, which we will release publicly to facilitate the evaluation of credibility and visual interpretability of the AQA system.
(4) Extensive experiments demonstrate that our FineParser achieves state-of-the-art performance with significant improvements and better visual interpretability.

\begin{figure*}[t]
  \centering
  \includegraphics[width=0.96\linewidth]{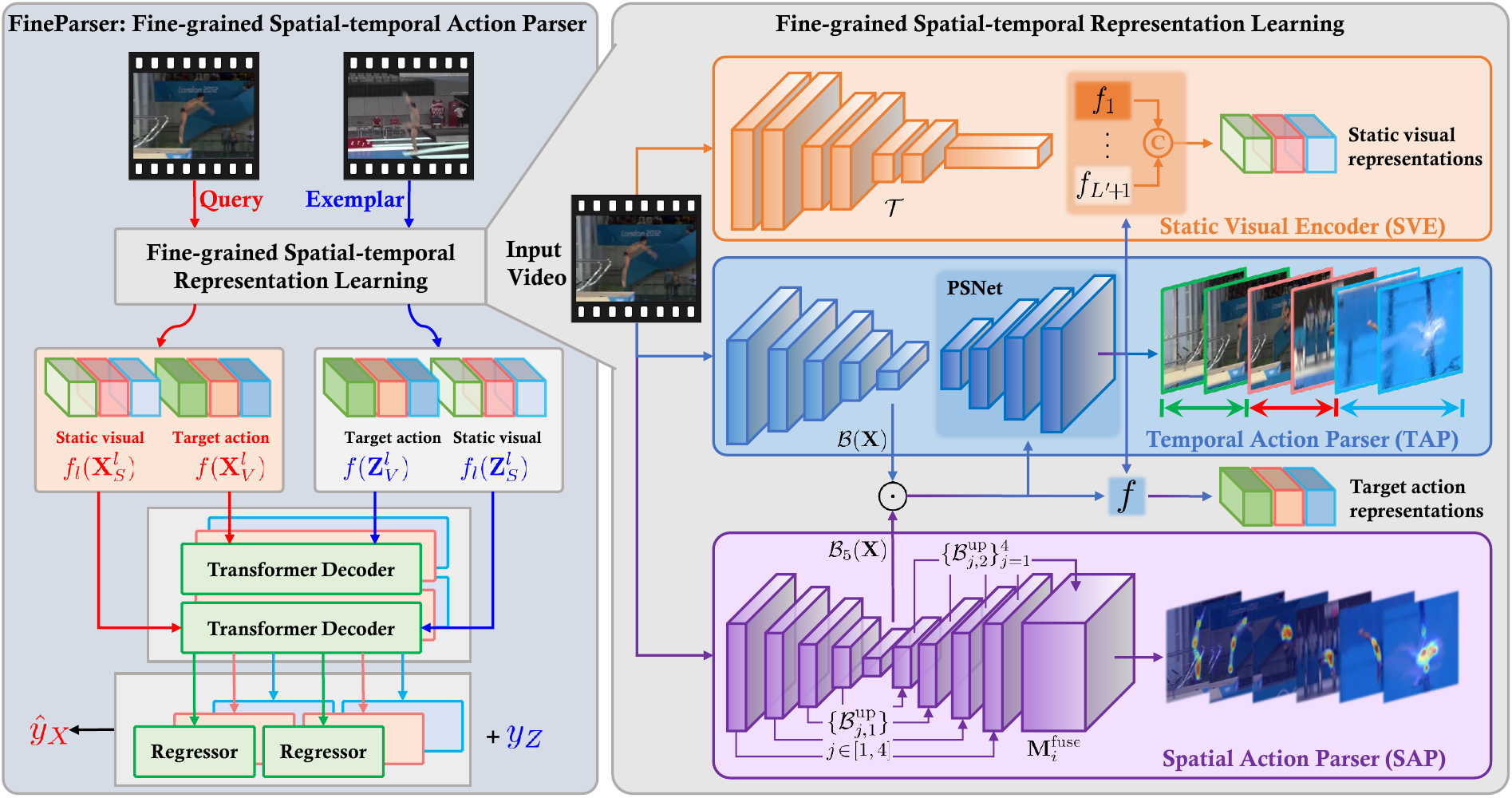}
  \vspace{-10pt}
  \caption{The architecture of the proposed \textit{\textbf{FineParser}}. Given a pair of query and exemplar videos, spatial action parser (SAP) and temporal action parser (TAP) extract spatial-temporal representations of human-centric foreground actions in pairwise videos, as well as predict both target action masks and step transitions. The static visual encoder (SVE) captures static visual representations combined with the target action representation to mine more contextual details. Finally, fine-grained contrastive regression (FineReg) utilizes the representations to predict the action score of the query video.}
  \label{pipeline}
\vspace{-9pt}
\end{figure*}
\section{Related Work}
\label{sec:relatedwork}

\textbf{Fine-grained Action Understanding.}
With ongoing advancements in action understanding, analyzing actions in finer granularity has become inevitable. Current endeavors in fine-grained action understanding mainly encompass tasks such as temporal action detection~\cite{piergiovanni2018fine,liu2021fineaction,gao2022fine}, action recognition \cite{zhu2018fine,munro2020multi,hong2021video}, video question answering \cite{cooray2020attention,zhang2021temporal,yu2023anetqa}, and video-text retrieval \cite{chen2020fine,doughty2022you}.
Recently, Shao~\textit{et al.}~\cite{shao2020finegym} constructed FineGym that provides coarse-to-fine annotations temporally and semantically for facilitating action recognition. Chen \textit{et al.}~\cite{chen2021sportscap} proposed SportsCap that estimates 3D joints and body meshes and predicts action labels. Li \textit{et al.}~\cite{li2021multisports} introduced MultiSports with spatio-temporal annotations of actions from four sports. Zhang \textit{et al.}~\cite{zhang2021temporal} constructed a temporal query network to answer fine-grained questions about event types and their attributes in untrimmed videos. Li \textit{et al.}~\cite{li2022weakly} presented a hierarchical atomic action network that models actions as combinations of reusable atomic ones to capture the commonality and individuality of actions. Zhang \textit{et al.}~\cite{zhang2023modeling} introduced a fine-grained video representation learning method to distinguish video processes and capture their temporal dynamics. These methods mainly concentrated on a fine-grained understanding of the temporal dimension. In contrast, our FineParser captures human-centric action representations by simultaneously building a fine-grained understanding in both time and space.

\noindent\textbf{Action Quality Assessment.}
In early pioneering work, Pirsiavash \textit{et al.} \cite{pirsiavash2014assessing} formulated the AQA task as a regression problem from action representations to scores, and Parisi \textit{et al.}~\cite{parisi2016human} adopted the correctness of performed action matches to assess action quality.
Parmar \textit{et al.}~\cite{parmar2017learning} demonstrated the effectiveness of spatio-temporal features for estimating scores in various competitive sports.
Recently, Tang \textit{et al.}~\cite{tang2020uncertainty} introduced an uncertainty-aware score distribution learning method to alleviate the ambiguity of judges' scores. Yu \textit{et al.}~\cite{yu2021group} developed a contrastive regression based on video-level features, enabling the ranking of videos and accurate score prediction. Wang \textit{et al.}~\cite{wang2021tsa} introduced TSA-Net to generate action representations using the outputs of the VOT tracker, improving AQA performance. Xu \textit{et al.}~\cite{xu2022finediving} contributed to a fine-grained sports video dataset for AQA and proposed a new action procedure-aware method to improve AQA performance.
Zhang \textit{et al.}~\cite{zhang2023logo} proposed a plug-and-play group-aware attention module to enrich clip-wise representations with contextual group information.
In contrast, our FineParser parses action in space and time to focus on the human-centric foreground action, improving AQA's credibility and visual interpretability.

\vspace{-3.5pt}
\section{Approach}
\label{approach}

This section presents a fine-grained spatial-temporal action parser for human-centric action quality assessment, i.e., \textit{\textbf{FineParser}}. As illustrated in \cref{pipeline}, FineParser consists of four components: spatial action parser (SAP), temporal action parser (TAP), static visual encoder (SVE), and fine-grained contrastive regression (FineReg).

\subsection{Problem Formulation}
Given a pair of query and exemplar videos with the same action type, denoted as $(\mathbf{X},\mathbf{Z})$, our approach is formulated as a fine-grained understanding framework that predicts the action score of the query video $\mathbf{X}$. Inspired by fine-grained contrastive regression~\cite{xu2022finediving}, our framework considers fine-grained quality differences between human-centric foreground actions in both time and space perspectives to model variations in their scores. The core is a new fine-grained action parser, FineParser $\mathcal{F}$, represented as
\begin{equation}
    \hat{y}_X=\mathcal{F}(\mathbf{X},\mathbf{Z},y_Z;\mathbf{\Theta}),
\end{equation}
where $\mathbf{\Theta}$ denotes all learnable parameters of $\mathcal{F}$, and $\hat{y}_X$ denotes the predicted action score of $\mathbf{X}$ referring to $\mathbf{Z}$ and its ground truth score $y_Z$.

\subsection{Fine-grained Spatio-temporal Action Parser}

FineParser is composed of four core components. In short, SAP, TAP, and SVE collaborate to learn fine-grained target action representations, and FineReg then uses these representations to predict the final score.

\noindent\textbf{Spatial Action Parser (SAP).} SAP parses the target action for each input video at a fine-grained spatial level. Inspired by I3D \cite{carreira2017quo} and its fully convolutional version \cite{munukutlaone}, transposed convolution layers are introduced before each max pooling layer to upsample the outputs of I3D submodules, and the rest after the last average pooling layer is discarded. These operations facilitate capturing multi-scale visual and semantic information that spans from short-term local features obtained from lower layers to long-term global semantic context derived from the last few layers.

Concretely, taking the query video $\mathbf{X}\!\!=\!\!\{\mathbf{X}_i\}_{i=1}^N$ as an example, the first I3D submodule $\mathcal{B}_1$ encodes each snippet $\mathbf{X}_i$ to capture short-term local features, as $\mathcal{B}_1(\mathbf{X})\!\!=\!\!\{\mathcal{B}_1(\mathbf{X}_i)\}_{i=1}^N$.
Similarly, other three submodules encode $\mathbf{X}_i$ to obtain middle representations, as $\mathcal{B}_j(\mathbf{X})\!\!=\!\!\mathcal{B}_j({\mathcal{B}_{j-1}(\mathbf{X})})$, with $j\!\!\in\!\![2, 4]$. For each $\mathcal{B}_j(\mathbf{X})$, two upsampling blocks are further inserted, denoted as $\mathcal{B}^{\text{up}}_{j,1}$ and $\mathcal{B}^{\text{up}}_{j,2}$. Both comprise convolution layers performed on the feature dimension and transpose convolution layers performed on both spatial and temporal dimensions. They can be presented as
\begin{align}
\mathbf{M}^{\text{up}_1}_{j,i}&=\mathcal{B}^{\text{up}}_{j,1}(\mathcal{B}_j(\mathbf{X}_i)),\ \mathbf{M}^{\text{up}_2}_{j,i}=\mathcal{B}^{\text{up}}_{j,2}(\mathcal{B}_j(\mathbf{X}_i)), \\
\mathbf{M}^{\text{fuse}}_i&=\text{Conv3d}(\text{Concat}(\{\mathbf{M}^{\text{up}_1}_{j,i}\}_{j=1}^4)),
\end{align}
where $\{\mathbf{M}^{\text{up}_2}_{j,i}\}_{j=1}^4$ are the predicted target action masks from different scales for optimizing SAP. These masks capture multi-scale human-centric foreground action information, from short-term local features obtained from lower layers (small scale) to long-term global semantic context derived from the last few layers (large scale).
$\mathbf{M}^{\text{fuse}}_i$ is the final target action mask of $\mathbf{X}_i$ by fusing $\{\mathbf{M}^{\text{up}_1}_{j,i}\}_{j=1}^4$.
SAP generates the above five target action masks and one target action mask embedding $\mathcal{B}_5(\mathbf{X})$, where the former are used to anticipate the human-centric foreground action mask, and the latter facilitates learning target action representations.
With mask embedding $\mathcal{B}_5(\mathbf{X})$ and video embedding $\mathcal{B}(\mathbf{X})$, target action representations $\mathbf{X}_{V}$ are calculated by elements-wise multiplication, as $\mathbf{X}_{V}\!\!=\!\!\mathcal{B}(\mathbf{X})\odot\text{sigmoid}(\mathcal{B}_5(\mathbf{X}))$.
For the exemplar video $\mathbf{Z}$, the target action representations $\mathbf{Z}_V$ can be obtained similarly.

\noindent\textbf{Temporal Action Parser (TAP).} TAP parses each action procedure into consecutive steps with semantic and temporal correspondences. Specifically, PSNet \cite{xu2022finediving} is adopted to parse $\mathbf{X}_{V}$ and $\mathbf{Z}_{V}$, which identifies the temporal transition when the step switches from one sub-action type to another. Supposed that $L'$ step transitions are needed to be identified in the action, the submodule $\mathcal{S}$ predicts the probability of the $k$-th step transiting at the $t$-th frame, denoted as $\mathcal{S}(\mathbf{X}_V)[t,k]\!\in\!\mathbf{R}$. By
\begin{align}
\hat{t}_k=\underset{\frac{T}{L'}(k\!-\!1)<\!t\leq\frac{T}{L'}k}{\arg\max}\mathcal{S}(\mathbf{X}_V)[t,k],
\end{align}
the timestamp $\hat{t}_k$ of the $k$-th step transition is predicted for each $k\!\in\![1, L']$. Based on $\{\hat{t}_k\}_{k=1}^{L'}$, each action procedure is parsed into $L'\!\!+\!\!1$ consecutive steps, i.e., $\{\mathbf{X}_V^l\}_{l=1}^{L'\!+\!1}$ and $\{\mathbf{Z}_V^l\}_{l=1}^{L'\!+\!1}$, where $l$ is the index of step.
While the lengths of the above consecutive steps may differ in nature, they are fixed to the same size via downsampling or upsampling operations $f$ along the temporal axis, ensuring that the dimensions of \textit{query} and \textit{key} are matched in the attention model. Therefore, the target action representations of query and exemplar videos become $\{f(\mathbf{X}_V^l)\}_{l=1}^{L'\!+\!1}$ and $\{f(\mathbf{Z}_V^l)\}_{l=1}^{L'\!+\!1}$.

\noindent\textbf{Static Visual Encoder (SVE).} 
SVE captures more contextual information to further enhance the action representations, especially for high-speed and complex actions like diving. It consists of two submodules: a ResNet model $\mathcal{T}$ and a set of projection functions $\{f_l\}_{l=1}^{L'\!+\!1}$. For the input video $\mathbf{X}$, the outputs of $\mathcal{T}$ can be obtained by
\begin{equation}
\begin{split}
    \mathbf{X}_S^1&=\mathcal{T}(\mathbf{X})[:\hat{t}_1],\ \mathbf{X}_S^{L'\!+\!1}=\mathcal{T}(\mathbf{X})[\hat{t}_{L'}:],\\ 
    \mathbf{X}_S^l&=\mathcal{T}(\mathbf{X})[\hat{t}_{l\!-\!1}:\hat{t}_l]~\text{s.t.}~ l\in[2, L'].
\end{split}
\end{equation}
Through post-projection, the static visual representations of $\mathbf{X}$ can be written as $\{f_l(\mathbf{X}_S^l)\}_{l=1}^{L'\!+\!1}$. Similarly, the static visual representation of $\mathbf{Z}$ are $\{f_l(\mathbf{Z}_S^l)\}_{l=1}^{L'\!+\!1}$.

\noindent\textbf{Fine-grained Contrastive Regression (FineReg).} It leverages the sequence-to-sequence representation ability of the transformer to learn powerful representations from pairwise steps and static visual representations via cross-attention. Specifically, the target action representations of pairwise steps $f(\mathbf{X}_V^l)$ and $f(\mathbf{Z}_V^l)$ interact with each other, helping the model focus on the consistent regions of motions in the cross-attention to generate the new features $\mathbf{D}_l^V$.
Similarly, cross-attention between the static visual representations of pairwise steps $f_l(\mathbf{X}_S^l)$ and $f_s(\mathbf{Z}_S^l)$ generates the new features $\mathbf{D}_l^S$.
Based on these two generated representations of the $l$-th step pairs, FineReg quantifies step quality differences between the query and exemplar by learning relative scores. This guides the framework to assess action quality at the fine-grained level with contrastive regression $\mathcal{R}$. The predicted score $\hat{y}_X$ of the query video $\mathbf{X}$ is calculated as
\begin{equation}
\hat{y}_X=\textstyle\sum_{l=1}^{L'\!+\!1}\lambda_l(\mathcal{R}_V(\mathbf{D}_l^V)+\mathcal{R}_S(\mathbf{D}_l^S)) + y_Z,
\end{equation}
where $\mathcal{R_V}$ and $\mathcal{R_S}$ are two three-layer MLPs with ReLU non-linearity, $y_Z$ is the ground truth score of the exemplar video $\mathbf{Z}$, and $\lambda_l$ is the coefficient weighting the relative score of the $l$-th step pairs.

\begin{figure*}[t]
  \centering
  \includegraphics[width=0.93\linewidth]{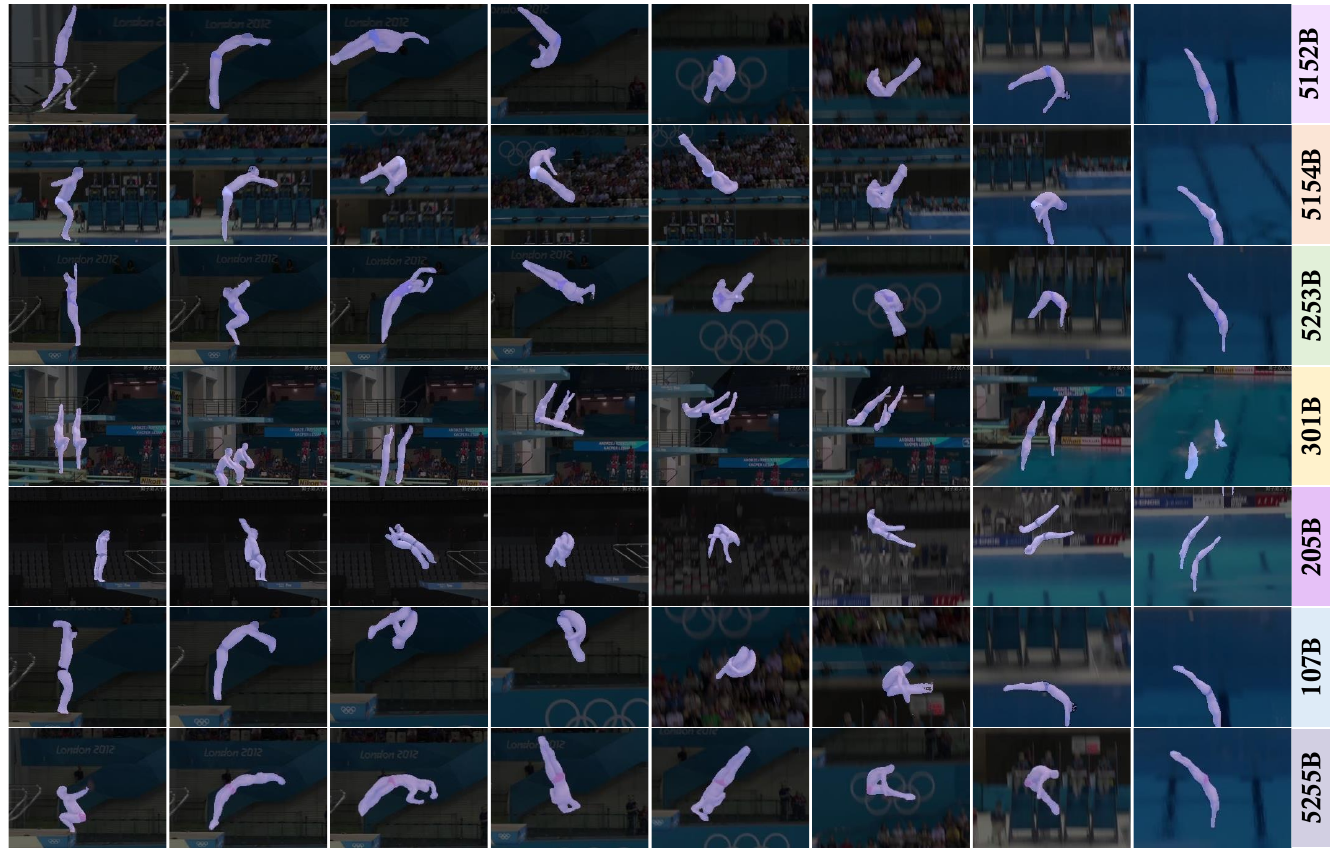}
  \vspace{-10pt}
  \caption{Examples of human-centric action mask annotations for the FineDiving dataset. The right line indicates the action type.}
  \label{mask_details}
\vspace{-10pt}
\end{figure*}

\subsection{Training and Inference}
\textbf{Training.} Given a pairwise query and exemplar videos $(\mathbf{X},\mathbf{Z})$ from the training set, FineParser is optimized by minimizing the following losses:
\begin{equation}
\mathcal{L}=\mathcal{L}_{\text{SAP}}+\mathcal{L}_{\text{TAP}}+\mathcal{L}_{\text{Reg}}.
\end{equation}
$\mathcal{L}_{\text{SAP}}$ is used to optimize SAP, calculated by
\begin{align}
\mathcal{L}_{\text{SAP}}\!&=\!\textstyle\sum\mathcal{L}_{\text{Focal}}(p(\mathrm{M}_{j,i})),\\
\mathcal{L}_{\text{Focal}}(p(\mathrm{M}_{j,i}))\!&=\!-\alpha_t(1-p(\mathrm{M}_{j,i}))^{\gamma}\log(p(\mathrm{M}_{j,i})),
\end{align}
where $\mathrm{M}_{j,i}\!\!=\!\!\mathrm{M}^{\text{up}_2}_{j,i}[l,h,w]$ is the element of $\mathbf{M}^{\text{up}_2}_{j,i}$, $p(\mathrm{M}_{j,i})\!\!=\!\!\mathrm{M}_{j,i}$ if the ground-truth mask $\mathrm{M}^{\text{gt}}_i\!=\!1$, and $p(\mathrm{M}_{j,i})\!\!=\!\!1\!-\!\mathrm{M}_{j,i}$, otherwise. $\mathcal{L}_{\text{Focal}}$ is the focal loss~\cite{Lin_2017_ICCV} between predicted and ground truth masks. $\mathcal{L}_{\text{TAP}}$ is used to optimize TAP, calculated by
\begin{align}
    \mathcal{L}_{\text{TAP}}\!\!=\!\!-\textstyle\!\!\sum_t(\mathrm{p}_k(t)\!\log\mathrm{S}_{t,k}\!+\!(1\!\!-\!\mathrm{p}_k(t))\log(1\!\!-\!\mathrm{S}_{t,k})),\!
\end{align}
where $\mathrm{S}_{t,k}\!\!=\!\!\mathcal{S}(\mathbf{X}_V)[t,k]$ is the predicted probability of the $k$-th step transiting at the $t$-th frame, and $\mathbf{p}_k$ is a binary distribution encoded by the ground truth timestamp $t_k$ of the $k$-th step transition, with $\mathrm{p}_k(t_k)\!\!=\!\!1$ and $\mathrm{p}_k(t_m)|_{m\neq k}\!\!=\!\!0$.
$\mathcal{L}_{\text{Reg}}$ is used to optimize $\mathcal{R}_V$ and $\mathcal{R}_S$ by minimizing the mean squared error between the ground truth $y_X$ and prediction $\hat{y}_X$, which is written as
\begin{equation}
    \mathcal{L}_{\text{Reg}}=\|\hat{y}_X-y_X\|^2.
\end{equation}

\noindent\textbf{Inference.} For a query video $\mathbf{X}$ from the testing set, the multi-exemplar voting strategy~\cite{yu2021group} is adopted to select $E$ exemplars $\{\mathbf{Z}_j)\}_{j=1}^E$ from the training set and construct pairwised $\{(\mathbf{X},\mathbf{Z}_j)\}_{j=1}^E$ with scores $\{y_{Z_j}\}_{j=1}^E$. The inference process can be written as
\begin{equation}
    \hat{y}_{X}=\frac{1}{E}\textstyle\sum_{j=1}^E(\mathcal{F}(\mathbf{X},\mathbf{Z}_j;\mathbf{\Theta})+y_{Z_j}).
\end{equation}

\section{Experiments}

\subsection{Datasets}
\begin{figure}[t]
  \centering
  \includegraphics[width=0.9\linewidth]{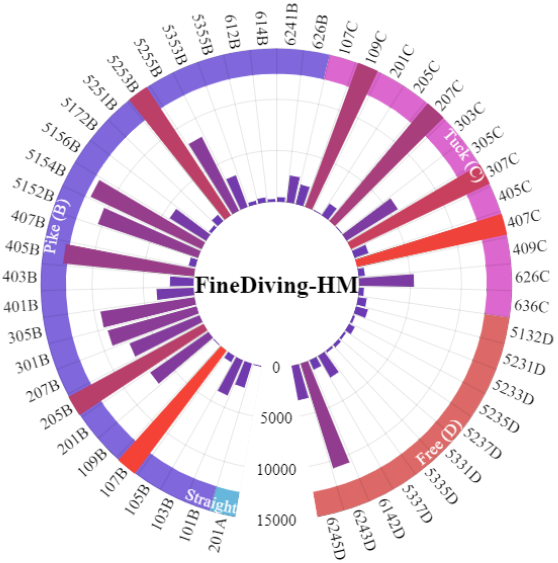}
  \vspace{-7pt}
  \caption{The distribution of human-centric foreground action masks. The largest number of mask instances is 35,287, belonging to the action type 107B. The smallest number of mask instances is 101, containing the action types 109B, 201A, 201C, and 303C.}
  \label{statistics}
\vspace{-9pt}
\end{figure}

\textbf{FineDiving-HM.}
FineDiving~\cite{xu2022finediving} contains 3,000 videos covering 52 action types, 29 sub-action types, 23 difficulty degree types, fine-grained temporal boundaries, and official action scores. To evaluate the effectiveness of our FineParser and make the results more credible and interpretable visually, we provide additional human-centric action mask annotations for the FineDiving dataset in this work, called \textbf{\textit{FineDiving-HM}}.
FineDiving-HM contains 312,256 mask frames covering 3,000 videos, in which each mask labels the target action region to distinguish the human-centric foreground and background.
FineDiving-HM mitigates the problem of requiring frame-level annotations to understand human-centric actions from fine-grained spatial and temporal levels. We employ three workers with prior diving knowledge to double-check the annotations to control their quality. \cref{mask_details} shows some examples of human-centric action mask annotations, which precisely focus on foreground target actions.
There are 312,256 foreground action masks in FineDiving-HM, where the number of action masks for individual diving is 248,713 and that for synchronized diving is 63,543. As shown in \cref{statistics}, the largest number of action masks is 35,287, belonging to the action type 107B; the second largest number of action masks is 34,054, belonging to the action type 407C; and the smallest number of action masks is 101, corresponding to the action types 109B, 201A, 201C, and 303C.
Coaches and athletes can use the above statistics to develop competition strategies, for instance, what led to the rise of 107B and 407C and how athletes gain a competitive edge.

\begin{table}[t]
  \centering
  \vspace{-5pt}
  \adjustbox{width=\linewidth}
  {
  \setlength{\tabcolsep}{13pt}
    \begin{tabular}{c|ccc}\toprule
        \multirow{2}{*}{Methods}  & \multicolumn{3}{c}{AQA Metrics}  \\
        \cmidrule(lr){2-4} 
        & $\rho\uparrow$ & \multicolumn{2}{c}{$R$-$\ell_2\downarrow(\times100)$} \\
        \midrule
        C3D-LSTM~\cite{parmar2017learning}   & 0.6969 & \multicolumn{2}{c}{1.0767} \\
        C3D-AVG~\cite{parmar2019and}  & 0.8371 & \multicolumn{2}{c}{0.6251}  \\
        MSCADC~\cite{parmar2019and} & 0.7688 & \multicolumn{2}{c}{0.9327}  \\
        I3D+MLP~\cite{tang2020uncertainty}  & 0.8776 & \multicolumn{2}{c}{0.4967} \\
        USDL~\cite{tang2020uncertainty}   & 0.8830  & \multicolumn{2}{c}{0.4800} \\
        MUSDL~\cite{tang2020uncertainty}   & 0.9241  & \multicolumn{2}{c}{0.3474} \\
        CoRe~\cite{yu2021group}  & 0.9308  & \multicolumn{2}{c}{0.3148} \\
        TSA~\cite{xu2022finediving}  & 0.9324 & \multicolumn{2}{c}{0.3022} \\
        \rowcolor{Gray} \textbf{FineParser} & \textbf{0.9435} & \multicolumn{2}{c}{\textbf{0.2602}} \\
        \midrule
        \multirow{2}{*}{Methods} & \multicolumn{3}{c}{TAP Metrics}  \\
        \cmidrule(lr){2-4} 
        & AIoU@0.5$\uparrow$  & \multicolumn{2}{c}{AIoU@0.75$\uparrow$} \\
        \midrule
        TSA~\cite{xu2022finediving}  & 0.9239 & \multicolumn{2}{c}{0.5007}  \\
        \rowcolor{Gray} \textbf{FineParser} & \textbf{0.9946} & \multicolumn{2}{c}{\textbf{0.9467}}  \\
        \midrule
        \multirow{2}{*}{Methods}  & \multicolumn{3}{c}{SAP Metrics}  \\
        \cmidrule(lr){2-4} 
        & MAE$\downarrow$ & $F_\beta\uparrow$ & $S_m\uparrow$ \\
        \midrule
        \rowcolor{Gray} \textbf{FineParser}  &\textbf{0.0408} & \textbf{0.1273} & \textbf{0.8357}  \\
        \bottomrule
    \end{tabular}
    }
    \vspace{-9pt}
    \caption{Comparisons of performance with state-of-the-art AQA methods on the FineDiving-HM Dataset. Our result is highlighted in the \textbf{bold} format. 
    }
  \label{ablution:sota}
  \vspace{-15pt}
\end{table}

\noindent\textbf{MTL-AQA.} It is a multi-task action quality assessment dataset~\cite{parmar2019and} consisting of 1,412 samples collected from 16 different world events, with annotations containing the degree of difficulty, scores from each judge (7 judges), type of diving action, and the final score.

\subsection{Evaluation Metrics}
\textbf{Action Quality Assessment.} 
Following previous efforts \cite{pan2019action,parmar2019and,tang2020uncertainty,yu2021group,xu2022finediving}, we utilize Spearman's rank correlation ($\rho$, the higher, the better) and Relative $\ell_2$ distance ($R_{\ell_2}$, the lower, the better) for evaluating the AQA task.

\noindent\textbf{Temporal Action Parsing.} Given the ground truth bounding boxes and a set of predicted temporal bounding boxes, we adopt the Average Intersection over Union (AIoU) \cite{xu2022finediving} to evaluate the performance of TAP. The higher the value of AIoU$@d$, the better the performance of TAP.

\noindent\textbf{Spatial Action Parsing.}
We adopt three evaluation metrics for comparison: MAE \cite{perazzi2012saliency}, F-measure $F_\beta$ ($\beta\!=\!0.3$) \cite{achanta2009frequency}, and S-measure $S_m$ \cite{fan2017structure}. MAE (the lower, the better) measures the average pixel-wise absolute error between the binary ground truth mask and normalized saliency prediction map. F-measure (the higher, the better) comprehensively considers precision and recall by computing the weighted harmonic mean. S-measure (the higher, the better) evaluates the structural similarity between the real-valued saliency map and the binary ground truth, considering object-aware ($S_o$) and region-aware ($S_r$) structure similarities ($\alpha\!=\!0.5$).

\subsection{Implementation Details}
We adopted the I3D model pre-trained on the Kinetics \cite{carreira2017quo} as the backbone of the SAP and TAP modules, where SAP is composed of $\{\mathcal{B}_j\}_{j=1}^5$ and $\{\mathcal{B}^{\text{up}}_{j,1},\mathcal{B}^{\text{up}}_{j,2}\}_{j=1}^4$ with the initial learning rate 10$^{-3}$ and TAP consists of $\mathcal{B}$ and $\mathcal{S}$ with the initial learning rate 10$^{-4}$. 
SAP and TAP did not share parameters.
Besides, we set the initial learning rates of $\mathcal{T}$ (i.e., ResNet34) in SVE as 10$^{-3}$. We utilized Adam \cite{kingma2014adam} optimizer and set weight decay as 0. In SAP and TAP, following previous works \cite{tang2020uncertainty,yu2021group,xu2022finediving}, we extracted 96 frames for each video and split them into 9 snippets, where each snippet contains 16 continuous frames with a stride of 10 frames.
We set $L'$ as 3 and the weights $\{\lambda_l\}_{l=1}^{L'}$ as $\{3,5,2\}$.
Furthermore, we followed the exemplar selection criterion in \cite{xu2022finediving} and \cite{yu2021group} on the FineDiving-HM and MTL-AQA datasets, respectively.
Following the experiment settings in \cite{tang2020uncertainty,yu2021group,xu2022finediving}, we selected 75 percent of samples for training and 25 percent for testing in all the experiments.

\subsection{Comparison with the State-of-the-Arts}
\textbf{FineDiving-HM.}
\cref{ablution:sota} summarized the experimental results of state-of-the-art AQA methods on the FineDiving-HM dataset. Our FineParser significantly improved the performance of Spearman's rank correlation and Relative $\ell_2$-distance compared to all methods. The advantages of FineParser stemmed from a fine-grained understanding of human-centric foreground actions, which requires the model to parse actions in time and space, making the model credible and interpretable visually.
Compared to C3D-LSTM, C3D-AVG, MSCADC, I3D+MLP, USDL, and MUSDL, FineParser outperformed them significantly and achieved 24.66\%, 10.64\%, 17.47\%, 6.59\%, 6.05\%, and 1.94\% performance improvements in terms of Spearman's rank correlation as well as 0.8165, 0.3649, 0.6725, 0.2365, 0.2198, and 0.0872 in Relative $\ell_2$-distance.
Compared to CoRe, FineParser obtained 1.27\% and 0.0546 performance improvements on Spearman's rank correlation and Relative $\ell_2$-distance. FineParser further improved the performance of TSA on Spearman's rank correlation and Relative $\ell_2$-distance, which also can be observed in the TAP metric.

\begin{table}[t]
\centering
\vspace{-5pt}
\adjustbox{width=\linewidth}
 {
  \setlength{\tabcolsep}{16pt}
    \begin{tabular}{c|cc}
    \toprule
    \multirow{2}{*}{Methods} 
    & \multicolumn{2}{c}{MTL-AQA} \\ 
    \cmidrule(lr){2-3}
    & $\rho\uparrow$ & $R$-$\ell_2\downarrow(\times100)$ \\
    \midrule
    Pose+DCT~\cite{pirsiavash2014assessing}   & 0.2682 & /  \\
    C3D-SVR~\cite{parmar2017learning}  & 0.7716 & / \\
    C3D-LSTM~\cite{parmar2017learning}  & 0.8489 & /  \\ 
    C3D-AVG-STL~\cite{parmar2019and}  & 0.8960 & / \\
    C3D-AVG-MTL~\cite{parmar2019and}  & 0.9044 & /  \\  
    USDL~\cite{tang2020uncertainty}  & 0.9231 & 0.4680 \\  
    MUSDL~\cite{tang2020uncertainty}  & 0.9273 & 0.4510  \\
    TSA-Net~\cite{wang2021tsa}  & 0.9422 & /  \\ 
    CoRe~\cite{yu2021group}  & 0.9512 & 0.2600  \\
    \rowcolor{Gray}\textbf{FineParser}  & \textbf{0.9585} & \textbf{0.2411} \\ 
    \bottomrule
    \end{tabular}
}
\vspace{-9pt}
\caption{\small Comparisons of performance with representative AQA methods on the MTL-AQA dataset. Our result is highlighted in the \textbf{bold} format.}
\label{tab:ablation-mtl-aqa}
\vspace{-15pt}
\end{table}

\noindent\textbf{MTL-AQA.}
\cref{tab:ablation-mtl-aqa} reported the experimental results of representative AQA methods on the MTL-AQA dataset. Our FineParser outperformed other methods on Spearman's rank correlation. For instance, FineParser achieved better AQA performance than CoRe and TSA-Net, demonstrating the effectiveness of additional human-centric foreground action masks and the meticulous design of a fine-grained action understanding of FineParser.

\begin{table}[t]
  \centering
  \vspace{-5pt}
  \adjustbox{width=\linewidth}
  {
  \setlength{\tabcolsep}{17pt}
    \begin{tabular}{c|ccccc}\toprule
        \multirow{2}{*}{Methods}
       & \multicolumn{3}{c}{Modules}  \\
        \cmidrule(lr){2-4} 
        & SAP  & SVE & TAP\\
        \midrule
        A & \Checkmark &   \\
        B &  & \Checkmark  \\
        C & & & \Checkmark  \\
        D & & \Checkmark & \Checkmark  \\
        E & \Checkmark & \Checkmark & \Checkmark \\
        \midrule
        Methods & {$\rho\uparrow$} & \multicolumn{2}{c}{{$R$-$\ell_2\downarrow(\times100)$}} \\
        \midrule
        A & 0.9313 & \multicolumn{2}{c}{0.3094} \\
        B & 0.9328 & \multicolumn{2}{c}{0.3097}  \\
        C & 0.9334 & \multicolumn{2}{c}{0.3122} \\
        D & 0.9351 & \multicolumn{2}{c}{0.2881} \\
        \rowcolor{Gray}E & \textbf{0.9435} & \multicolumn{2}{c}{\textbf{0.2602}} \\
        \midrule
        
        \multirow{2}{*}{Methods}
        & \multicolumn{3}{c}{TAP Metrics}  \\
        \cmidrule(lr){2-4} 
        & AIoU@0.5$\uparrow$  & \multicolumn{2}{c}{AIoU@@0.75$\uparrow$} \\
        \midrule
        C & 0.9907& \multicolumn{2}{c}{0.9039}  \\
        D & 0.9920& \multicolumn{2}{c}{0.8932}  \\
        \rowcolor{Gray}E & \textbf{0.9946}& \multicolumn{2}{c}{\textbf{0.9467}}  \\
        \midrule
        \multirow{2}{*}{Methods}
        & \multicolumn{3}{c}{SAP Metrics}  \\
        \cmidrule(lr){2-4}
        & MAE$\downarrow$ & $F_\beta\uparrow$ & $S_m\uparrow$ \\
        \midrule
        \rowcolor{Gray}E &  \textbf{0.0408} & \textbf{0.1273} & \textbf{0.8357} \\
        \bottomrule
    \end{tabular}
    }
    \vspace{-9pt}
    \caption{Ablation study on different modules in FineParser on FineDiving-HM. The results of unavailable methods are omitted.}
  \label{ablution:modules}
  \vspace{-11pt}
\end{table}

\begin{figure*}[t]
    \centering
    \includegraphics[width=0.99\textwidth]{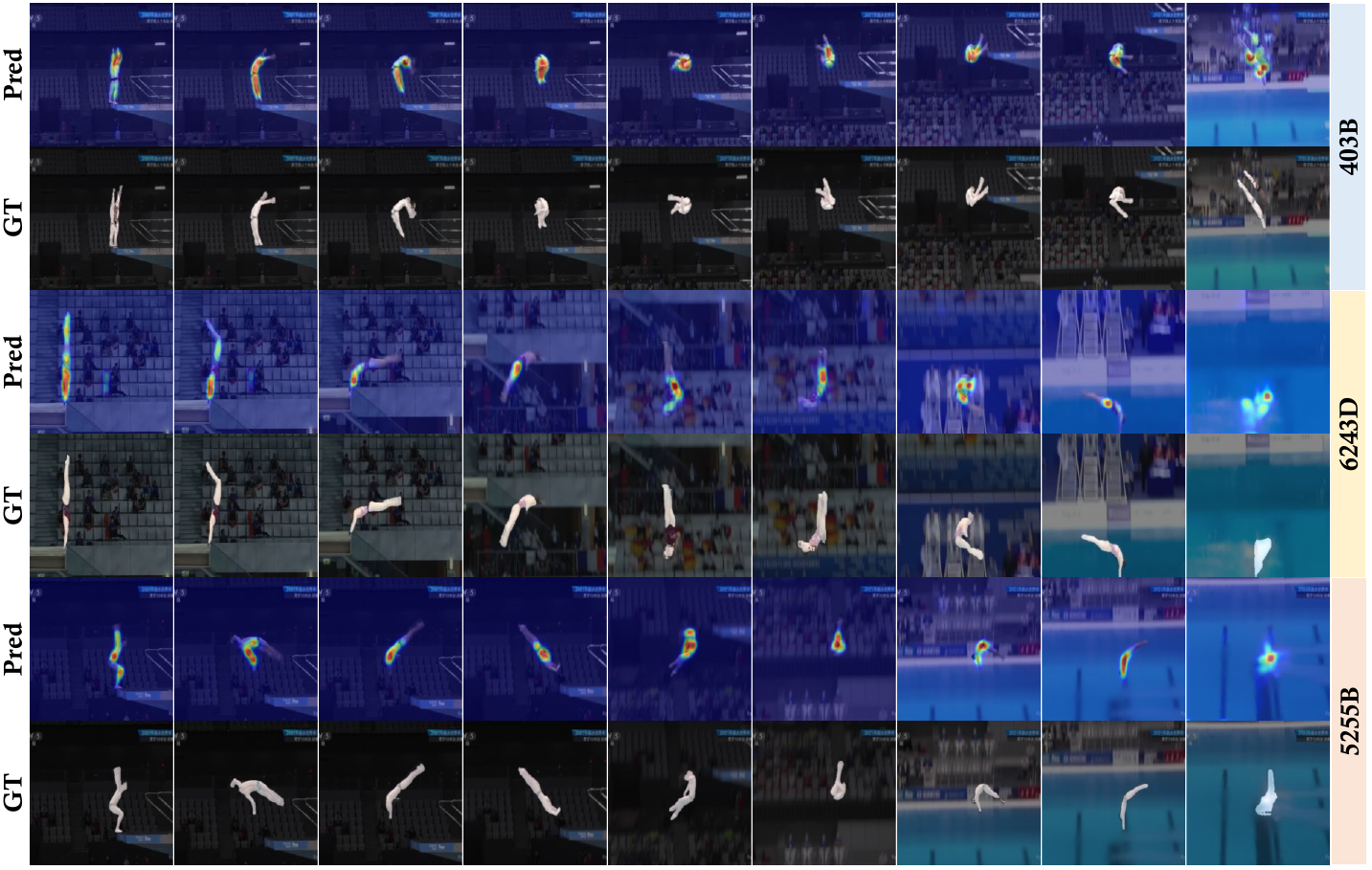}
    \vspace{-10pt}
    \caption{Visualization of the predictions of target action masks produced by SAP. The predicted masks can focus on the target action regions in each frame, minimizing the impact of invalid backgrounds on action quality assessment.}
    \vspace{-7pt}
    \label{mask_pred}
\end{figure*}

\subsection{Ablation Study}
We conducted an ablation study on the FineDiving-HM dataset to demonstrate the effectiveness of individual parts of FineParser by designing different modules, different backbones of SVE, and varied step durations of the projection function in SVE.

\noindent\textbf{Different Modules in FineParser.}
We summarized the experimental results in \cref{ablution:modules}. Under Spearman's rank correlation, the AQA performance of the model with SVE and TAP can be improved from 0.9334 to 0.9351. Significant improvements on AIoU@0.5 and AIoU@0.75 are directly proportional to the accuracy of action quality assessment, demonstrating that SVE can help the model perform more accurate temporal action parsing in the TAP module. Further introducing the SAP module into the model, the AQA performance can be further enhanced to 0.9435 in Spearman's rank correlation, demonstrating that incorporating SAP allows for capturing more characteristics of target action, achieving more accurate action quality assessment.
If only SAP or SVE were introduced, Spearman's rank correlations would be 0.9313 or 0.9328, respectively, which cannot achieve the AQA performance of our final version.

\noindent\textbf{Different Step Durations in SVE.}
We studied the influence of different step durations used in the projection function of SVE on the AQA performance. As shown in \cref{ablution:duration}, we set the step duration as 2, 4, and 8 and then observe that the AQA performance of FineParser is optimal when set to 4. It is attributed to proper step duration that can benefit mining more valuable information from human-centric foreground action and static visual representations.

\noindent\textbf{Different Backbones of SVE.}
We conducted several experiments on the FineDiving-HM dataset to investigate the effects of different backbones of SVE on the performance of action quality assessment. In \cref{ablution:backbone}, ResNet34 outperforms other ResNet architectures while slightly inferior to ViT-S/16. For one thing, ResNet34 has a deeper network depth than ResNet18, allowing it to capture more global and high-level semantic information, whereas ResNet50 may lead to overfitting on the steps with relatively short durations (e.g., four frames). In addition, ViT allows the model to capture long-term dependencies among video frames rather than local relationships, which is beneficial to learning target action representations by capturing global features, further improving the AQA performance (i.e., $R$-$\ell_2$) of FineParser.

\subsection{Visualization}
To intuitively understand the benefits of our FineParser, we visualize the predicted masks obtained by SAP, as shown in \cref{mask_pred}. We see that the predictions can focus on target action regions in each frame, minimizing the impact of invalid backgrounds on action quality assessment.

\begin{table}[t]
  \centering
  \adjustbox{width=\linewidth}
  {
  \setlength{\tabcolsep}{17pt}

    \begin{tabular}{c|ccccc}\toprule

        \multirow{2}{*}{Duration}
       & \multicolumn{3}{c}{AQA}  \\
        \cmidrule(lr){2-4}
        & {$\rho\uparrow$} & \multicolumn{2}{c}{{$R$-$\ell_2\downarrow(\times100)$}} \\
        \midrule
        2 & 0.9320 & \multicolumn{2}{c}{0.2994} \\
        \textbf{4} & \textbf{0.9435} & \multicolumn{2}{c}{\textbf{0.2602}}  \\
        8 & 0.9337 & \multicolumn{2}{c}{0.2940} \\
        \bottomrule
        
        \multirow{2}{*}{Duration}
       & \multicolumn{3}{c}{TAP}  \\
        \cmidrule(lr){2-4} 
        & AIoU@0.5$\uparrow$  & \multicolumn{2}{c}{AIoU@0.75$\uparrow$} \\
        \midrule
        2 & 0.9987 & \multicolumn{2}{c}{0.9359}  \\
        \textbf{4} & \textbf{0.9946} & \multicolumn{2}{c}{\textbf{0.9467}}  \\
        8 & 0.9973& \multicolumn{2}{c}{0.9493}  \\
        \midrule
        \multirow{2}{*}{Duration}
        & \multicolumn{3}{c}{SAP}  \\
        \cmidrule(lr){2-4}
        & MAE$\downarrow$ & $F_\beta\uparrow$ & $S_m\uparrow$ \\
        \midrule
        2 & 0.0532  & 0.1010 & 0.8643  \\
        \textbf{4} &  \textbf{0.0408}& \textbf{0.1273} & \textbf{0.8357}  \\
        8 &  0.0535 & 0.1057 & 0.8616 \\
        \midrule
    \end{tabular}
    }
    \vspace{-9pt}
    \caption{Ablation study on different step durations in the projection function in the SVE module.}
  \label{ablution:duration}
  \vspace{-15pt}
\end{table}

\begin{table}[t]
  \centering
  \adjustbox{width=\linewidth}
  {
  \setlength{\tabcolsep}{25pt}
    \begin{tabular}{c|cc}\toprule
        Backbones & {$\rho\uparrow$} & {$R$-$\ell_2\downarrow(\times100)$} \\
        \midrule
        ResNet18 & 0.9363 & 0.2829 \\
        \textbf{ResNet34} & \textbf{0.9435} & \textbf{0.2602} \\
        ResNet50 & 0.9362 & 0.2859 \\
        ViT-S/16 & 0.9426 & \underline{0.2583} \\
        \bottomrule
    \end{tabular}
    }
    \vspace{-7pt}
    \caption{Ablation study on different backbones in SVE.}
  \label{ablution:backbone}
  \vspace{-9pt}
\end{table}

\section{Conclusion and Discussion}

We presented an end-to-end fine-grained spatial-temporal action parser named FineParser for the AQA task. It learned fine-grained representations for target actions via integrating spatial action parser, temporal action parser, static visual encoder, and fine-grained contrastive regression and achieved state-of-the-art. To understand human-centric actions from fine-grained spatial and temporal levels, we also provided human-centric foreground action mask annotations for the FineDiving dataset, named FineDiving-HM, to provide three quantitative metrics for the credibility and visual interpretability of the AQA model. We hope FineParser could be a baseline for fine-grained human-centric AQA and facilitate more tasks that require a fine-grained understanding of sports.

\noindent\textbf{Limitations.} The human-centric foreground action masks need to be manually adjusted and labeled. This work contributes new human-centric annotations for the dataset on diving events, while they are challenging to transfer to other competitive sports directly.

{
    \small
    \bibliographystyle{ieeenat_fullname}
    \bibliography{main}
}

\end{document}